%% file: main.tex
 \documentclass[pmlr,twocolumn,10pt]{jmlr} 

\usepackage{booktabs}
\usepackage{siunitx}

\usepackage[switch]{lineno}

\usepackage{caption}
\usepackage{graphicx}
\usepackage{subcaption}
\usepackage{array}
\usepackage{ragged2e}
\usepackage{caption}
\usepackage{multirow}
\usepackage{colortbl}
\usepackage{amsmath}
\usepackage{caption}
\usepackage{xcolor}
\usepackage{array}
\usepackage{xspace}
\usepackage{ragged2e}
\usepackage{colortbl}
\usepackage{booktabs}
\usepackage{subcaption}
\usepackage{tikz}
\usepackage{bbding}

\usepackage[most,skins,theorems]{tcolorbox}

\tcbset{
  aibox/.style={
    width=\linewidth,
    top=8pt,
    bottom=4pt,
    colback=blue!6!white,
    colframe=black,
    colbacktitle=black,
    enhanced,
    center,
    attach boxed title to top left={yshift=-0.1in,xshift=0.05in},
    boxed title style={boxrule=0pt,colframe=white,},
  }
}

\newtcolorbox{AIbox}[2][]{aibox,title=#2,#1}

\definecolor{yellowtext}{RGB}{68,132,243}
\definecolor{yellowred}{RGB}{50,167,82}
\definecolor{yellowblue}{RGB}{251,191,5}

\newcommand{\TextCircle}[1][0.7]{%
    \tikz[baseline=(char.base)]\node[shape=circle,draw=black,inner sep=0pt,line width=0.1pt,minimum size=1.1em,fill=yellowtext,text=white,scale=#1] (char) {T};
}
\newcommand{\TextImage}[1][0.7]{%
    \tikz[baseline=(char.base)]\node[shape=circle,draw=black,inner sep=0pt,line width=0.1pt,minimum size=1.1em,fill=yellowred,text=white,scale=#1] (char) {I};
}

\newcommand{\ours}{\textsc{MedVLThinker}\xspace}

\theorembodyfont{\upshape}
\theoremheaderfont{\scshape}
\theorempostheader{:}
\theoremsep{\newline}

\jmlrvolume{297}
\jmlryear{2025}
\jmlrworkshop{Machine Learning for Health (ML4H) 2025} 

 \title[\ours]{\ours: Simple Baselines for Multimodal Medical Reasoning}

 \author{%
  \Name{Xiaoke Huang} \Email{xhuan192@ucsc.edu}\\
  \addr UC Santa Cruz \\
  \Name{Juncheng Wu} \Email{jwu418@ucsc.edu}\\
  \addr UC Santa Cruz \\
  \Name{Hui Liu} \Email{huiliulayne@gmail.com}\\
  \addr Amazon Research \\
  \Name{Xianfeng Tang} \Email{tangxianfeng@outlook.com}\\
  \addr Amazon Research \\
  \Name{Yuyin Zhou} \Email{yzhou284@ucsc.edu}\\
  \addr UC Santa Cruz
 }

\begin{document}

\maketitle

\input{sections/0_abstract}
\begin{keywords}
Medical, Reasoning, Multimodal, Vision-Language, Health Care
\end{keywords}

\paragraph*{Data and Code Availability}
Our code, models, and data are publicly available at~\url{https://github.com/UCSC-VLAA/MedVLThinker}.

\paragraph*{Institutional Review Board (IRB)}
Our research does not require IRB approval.

\begin{figure}[!t]
  \centering
  
    \includegraphics[width=0.7\linewidth]{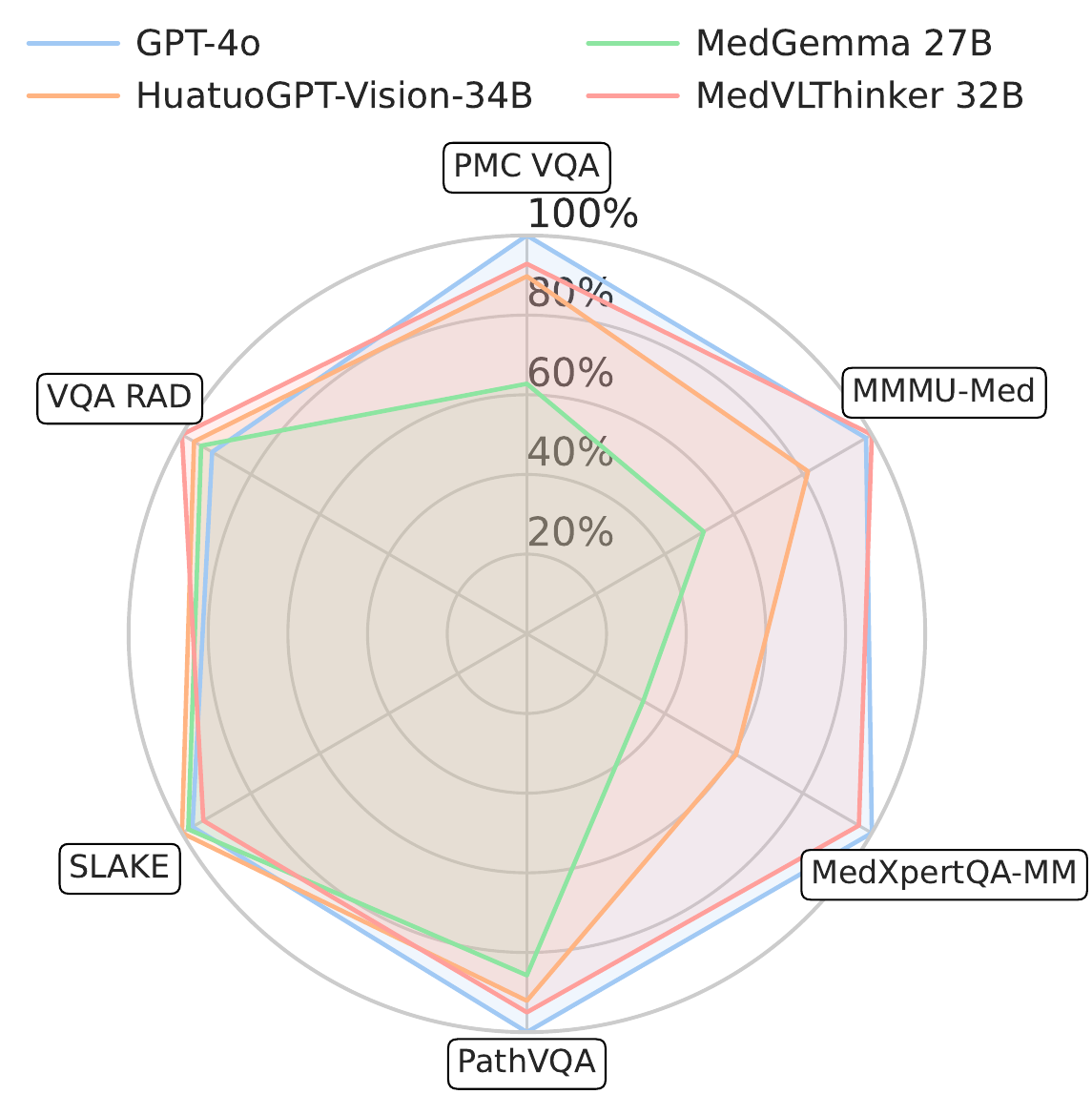}
  
    \caption{
        \ours~provides a simple yet strong baseline for multimodal medical reasoning. Notably, \ours-32B yields performance \textbf{on par with the closed-source GPT-4o model}.
    }
    \label{fig:radar}
\end{figure}

\begin{figure*}[t]
    \centering
    \makebox[\linewidth][c]{
        \includegraphics[width=1.00\linewidth]{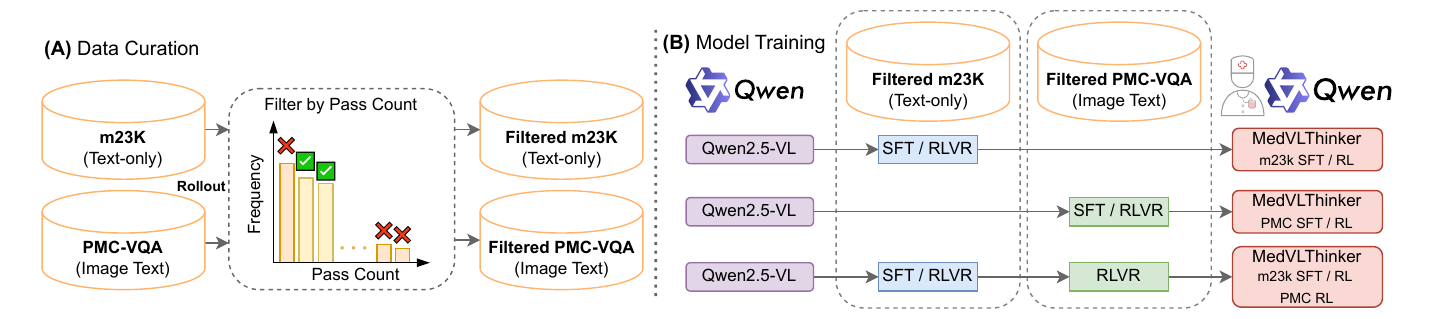}
    }
    \caption{
    The data filtering and training pipeline. 
    (A) We first filter both text-only m23k dataset and image-text PMC-VQA dataset, by generating multiple answers per question with Qwen2.5-VL-Instruct. Then we filter those questions are answered all wrong or almost correct.
    (B) Based on the filtered two datasets, we conduct supervised finetuning (SFT), reinforcement learning with verfiable rewaresd (RLVR), and their combination to train a herd of multimodal medical large reasoning models.
    }
    \label{fig:method}
\end{figure*}

\input{sections/1_intro}

\input{sections/2_related_works}

\input{sections/3_method}

\input{tabs/comp}

\input{tabs/comp-others}

\input{sections/4_experiments}

\input{sections/5_discussions}

\acks{This work was partially funded by an unrestricted gift from Google.}

\bibliography{main}

\appendix

\input{sections/6_supp}

\end{document}

%% file: sections/0_abstract.tex
\begin{abstract}

Large Reasoning Models (LRMs) have introduced a new paradigm in AI by enabling models to ``think before responding" via chain-of-thought reasoning.  However, the absence of open and reproducible recipes for building reasoning-centric medical LMMs hinders community-wide research, analysis, and comparison. In this paper, we present \ours, a suite of simple yet strong baselines. Our fully open recipe consists of: (1) systematic data curation for both text-only and image-text medical data, filtered according to varying levels of reasoning difficulty, and (2) two training paradigms: Supervised Fine-Tuning (SFT) on distilled reasoning traces and Reinforcement Learning with Verifiable Rewards (RLVR) based on final answer correctness. Across extensive experiments on the Qwen2.5-VL model family (3B, 7B) and six medical QA benchmarks, we find that RLVR consistently and significantly outperforms SFT. Additionally, under the RLVR framework, a key, counter-intuitive finding is that training on our curated text-only reasoning data provides a more substantial performance boost than training on multimodal image-text data.
Our best open 7B model, trained using the RLVR recipe on text-only data, establishes a new state-of-the-art on existing public VQA benchmarks, surpassing all previous open-source medical LMMs. Furthermore, scaling our model to 32B achieves performance on par with the proprietary GPT-4o.
We release all curated data, models, and code to provide the community with a strong, open foundation for future research in multimodal medical reasoning.

\end{abstract}

%% file: sections/1_intro.tex
\section{Introduction}

The practice of healthcare increasingly involves processing vast amounts of multimodal medical data (e.g., text, imaging, lab results). Clinicians must integrate information from different sources (clinical notes, radiology images, lab reports) to make diagnoses and treatment decisions.  Large Multimodal Models (LMMs) have recently emerged as general-purpose foundation models that can perceive and reason about visual inputs~\cite{li2023llava,liu2023visual,liu2024improved,hurst2024gpt,chen2024huatuogpt,xie2024medtrinity}. Given that medical data are natively multimodal (e.g., microscopy slides, CT and MRI scans, X-rays), LMMs have a natural appeal for medical AI and have begun to be adopted in modality-rich clinical settings with the potential to improve diagnosis~\cite{chen2024huatuogpt,li2023llava,liu2023medical}, treatment planning~\cite{zhou2023skingpt}, and patient monitoring~\cite{alshibli2025vision}.

Parallel to this, Large Reasoning Models (LRMs) extend large language models with a new response paradigm: the model “thinks” through a chain-of-thought before producing a final answer. This allows the model to devote more computation at inference time to reasoning, often improving performance on complex tasks~\cite{guo2025deepseek,guha2025openthoughts,jaech2024openai}. Early medical adaptations of text-only LRMs have demonstrated strong performance on medical QA tasks~\cite{huang2025m1,chen2024huatuogpto1,wu2025medreason,jiang2025meds,xie2024preliminary}. The ability to generate detailed reasoning steps at test time appears to confer significant gains in accuracy on challenging questions~\cite{zuo2025medxpertqa}. However, how to best combine this reasoning paradigm with multimodal understanding remains underexplored. While there have been efforts to build medical multimodal reasoning models,  they are often limited in openness—being either entirely closed-source~\cite{su2025gmai,liu2025x}, releasing only model weights without data or training code~\cite{sellergren2025medgemma}, or, if fully open, are confined to narrow datasets or specific domains (e.g., CT or MRI only)~\cite{lai2025med,pan2025medvlm}. As a consequence, the field lacks a comprehensive analysis of how critical factors such as data modality, curation pipelines, and training strategies affect model performance.



In this paper, we provide \ours, the very first fully open-source recipe for building and evaluating generalized Medical Vision-Language Reasoning Models. Our comprehensive framework provides a complete workflow, from data curation and training pipelines to a standardized evaluation protocol. This enables, for the first time, \emph{a fair and systematic comparison across diverse multimodal medical QA benchmarks}.
Figure~\ref{fig:method} provides an overview of our approach. We first curate \textbf{two types of training data}: a \textit{text-only} QA dataset and an \textit{image-text} (multimodal) QA dataset. Using a general-purpose multimodal LLM (Qwen2.5-VL-Instruct)~\cite{bai2025qwen2}, we probe each question with multiple trials to estimate its difficulty. Specifically, for each question, we generate multiple candidate answers and count how many times the model answers correctly (the “pass count”). Questions that are consistently answered correctly (too easy) or never answered correctly (too hard) are filtered out, yielding a focused training set of medium-difficulty questions. We then employ strong teacher models to generate detailed reasoning chains (long \textit{chains-of-thought}, CoTs) for the remaining questions. For text-only questions, we use the \textit{DeepSeek}~\cite{guo2025deepseek} model (a powerful text-based LRM) as the CoT teacher, and for image-based questions, we use GPT-4o~\cite{hurst2024gpt} (a vision-enabled GPT-4 variant).

Using these data, we train the base multimodal LLM under \textbf{two paradigms}: (1) Supervised fine-tuning (SFT) on the teacher-generated CoT traces, and (2) Reinforcement Learning with Verifiable Rewards (RLVR) on the question-answer pairs (without CoTs). SFT directly teaches the model to reproduce high-quality reasoning and answer traces, whereas RLVR uses only binary rewards from answer correctness to encourage the model’s own reasoning. We implement RLVR via Group Relative Policy Optimization (GRPO)~\cite{shao2024deepseekmath}, an efficient policy-gradient algorithm that requires no value estimator or critic model. In RLVR training, the model generates multiple reasoning traces for each question; each trace is verified for correct answer format and correctness of the final answer, yielding a +1 or -1 reward. These binary rewards are normalized (whitened) across the batch and fed into the GRPO update step, which applies a PPO-style clipped objective. This process gradually concentrates the model’s generation probability mass on verifiably correct reasoning traces while limiting divergence from the original model output distribution.

We conduct extensive experiments on \textbf{six multimodal medical QA benchmarks} to investigate the properties of our \ours. We use the Qwen2.5-VL series as the base models (in 3B, 7B, and 32B parameter sizes). Our evaluations cover both general medical visual QA and modality-specific QA (covering specialties like pathology, radiology, etc.). The results reveal several important, and at times counter-intuitive, findings:
First, regarding training paradigms, models trained with RLVR consistently outperform those trained with SFT across both 3B and 7B scales.
Second, in terms of data modality, text-only training outperforms image-text training. Notably, SFT on distilled text-only CoT data degrades performance relative to the base model (e.g., \ours-7B accuracy drops from 53.5\% to 43.8\%), whereas SFT on image-text data yields performance similar to the untrained base model. In contrast, RLVR on text-only data provides the largest performance boost, improving the 7B model from 53.5\% to 54.9\%. RLVR on image-text data also improves performance, but to a lesser extent. Moreover, combining text, only and image-text data—either through SFT+RL or sequential RL, does not yield additional gains beyond using text-only data alone.
Third, model scale has a clear impact: 7B models consistently outperform their 3B counterparts across all configurations. 

Among existing open-source 7B medical LMMs, \ours-7B (trained with RLVR on text-only data) achieves a new state-of-the-art average accuracy of 54.9\% across six benchmarks. To evaluate the effect of model scaling, we further train a 32B variant. As shown in Figure~\ref{fig:radar}, \ours-32B performs competitively with the proprietary GPT-4o, demonstrating the potential of open models to close the performance gap with commercial systems. To accelerate community-driven development and foster future innovation, we will release our complete research toolkit, including all models, code, and pipelines for data curation, training, and evaluation.

%% file: sections/2_related_works.tex
\section{Related Works}

\subsection{Large Reasoning Models and Medical Adaptation}
Large Reasoning Models (LRMs) endow large language models with the ability to articulate step-by-step reasoning before finalizing an answer~\cite{wei2022chain,guo2025deepseek,team2025kimi,jaech2024openai}. This test-time “think then answer” approach allows extended reasoning and has yielded impressive gains in domains such as mathematical problem~\cite{zeng2025simplerl,yang2025qwen3,muennighoff2025s1} solving and code generation~\cite{jaech2024openai,yang2024chain}. One way to train LRMs is via Reinforcement Learning with Verifiable Rewards (RLVR), which forgoes supervised chain-of-thought data and instead uses binary feedback on answer correctness~\cite{chen2025minimax,yu2025dapo}. RLVR eliminates the need to curate lengthy reasoning exemplars; it directly incentivizes correct reasoning by rewarding only the final outcome. In practice, an efficient implementation of RLVR is crucial. Group Relative Policy Optimization (GRPO)~\cite{shao2024deepseekmath} has been adopted for its efficiency, removing the need for a separate value network (critic) during RL updates. An alternative approach is to distill the reasoning traces of stronger models via supervised fine-tuning (SFT). For example, one can use a GPT-4 level model to generate high-quality explanations (CoTs) for medical questions, and then fine-tune a smaller model on this data~\cite{chen2024huatuogpto1}. Recent work shows that fine-tuning medical-focused LRMs (either via SFT on expert traces or via RL on answer rewards) can significantly improve medical question answering performance~\cite{wu2025medreason,huang2025m1,jiang2025meds}. Our work extends these ideas to the multimodal realm, examining whether similar reasoning enhancements hold when visual information is involved.

\subsection{Multimodal Medical Large Language Models}
Given that clinical data often includes images (radiology~\cite{lau2018dataset}, pathology~\cite{ikezogwo2023quilt}, etc.), there is growing interest in extending LLMs to handle visual inputs for medical applications. Med-Flamingo~\cite{moor2023med} was among the first to propose an interleaved vision-language training pipeline for a medical LLM, enabling it to handle image-text pairs in a single prompt. LLaVA-Med~\cite{li2023llava} introduced a two-stage approach: first, connect a vision encoder with an LLM via a learned projection (connector) and fine-tune on general images; second, fine-tune the combined model on medical image–text instruction data to specialize it. \textit{PMC-VQA}~\cite{zhang2023pmc} is one such large-scale multimodal instruction dataset, constructed from PubMed Central articles (figures and captions) using GPT-3.5 as an annotator. However, the quality of GPT-3.5-generated questions and answers in PMC-VQA is limited by the base model’s capacity, and the dataset likely contains noise or insufficiently detailed questions. Other contemporaneous efforts include HuatuoGPT-Vision~\cite{chen2024huatuogpt}, which scales up LLaVA-Med’s pipeline by generating a much larger set of QA pairs from a medical corpus and training larger models (up to 34B parameters). There are also modality-specific medical VLMs such as RadFM~\cite{wu2023towards} and SkinGPT~\cite{zhou2024pre} that follow similar pipelines but focus on particular domains (e.g., radiology, dermatology) with domain-specific image-text data. In summary, several open-source medical LMMs have been proposed, but integrating an explicit reasoning mechanism (as in LRMs) into these models has not been thoroughly studied prior to our work.

\subsection{Concurrent Works}
Very recently, a few works have begun exploring the idea of eliciting \emph{medical reasoning} in LLMs. For text-only medical QA, HuatuoGPT-o1 uses a PPO-based RL approach~\cite{schulman2017proximal} with an external reward model to train a medical reasoning LLM~\cite{chen2024huatuogpto1}, and MedS3 leverages Process-Reward Models (PRMs) for RL to improve stepwise reasoning~\cite{jiang2025meds}. Another approach, denoted M1 in a recent preprint, distills the reasoning traces of a GPT-4-based model (denoted R1) into a smaller model via SFT~\cite{huang2025m1}. In the multimodal domain, MedVLM-R1~\cite{pan2025medvlm} demonstrates the effectiveness of RLVR on a small scale of multimodal data (fewer than 1K training samples), and Med-R1~\cite{lai2025med} applies a similar RLVR scheme on separate modality-specific datasets. However, these models are trained on limited data and are not generalizable across different types of medical visual questions. GMAI-VL-R1~\cite{su2025gmai} is a general multimodal medical LLM trained with an RLVR paradigm, but its training data and code are not publicly available. In contrast, our work provides an \textbf{open-source} recipe for building multimodal medical reasoning models with both SFT and RL techniques, and we conduct a thorough experimental study across varying model scales (3B, 7B, 32B) and diverse benchmarks. To our knowledge, this is the first work to systematically compare supervised CoT distillation and RLVR for multimodal medical QA, and to benchmark the resulting models against prior open medical LMMs and closed models like GPT-4.

%% file: sections/3_method.tex
\section{Methods}
\label{sec3:method}
We describe our data curation process and training methodologies for \ours. Figure~\ref{fig:method} illustrates the overall pipeline of data filtering and model training.

\textbf{Data Curation and Filtering.} We gather two datasets for training: a text-only medical QA dataset and a multimodal (image+text) medical QA dataset. For text-only data, we use the \texttt{m23k}~\cite{huang2025m1}, which compiles 23,493 multiple-choice medical questions from the training splits of MedQA~\cite{jin2021disease}, MedMCQA~\cite{pal2022medmcqa}, and HeadQA~\cite{vilares2019head}. Each question in m23k is accompanied by a set of candidate answers, and we have access to high-quality reasoning chains (CoTs) for these questions distilled from the DeepSeek-R1~\cite{guo2025deepseek} model. For multimodal data, we adopt PMC-VQA~\cite{zhang2023pmc}, a large dataset of 176,948 visual QA pairs derived from biomedical literature figures and captions (covering about 149k images). PMC-VQA was generated using GPT-3.5 and covers a broad range of medical topics, making it a general-purpose multimodal medical QA resource (unlike modality-specific datasets such as PathVQA~\cite{he2020pathvqa}, SLAKE~\cite{liu2021slake}, VQA-Rad~\cite{lau2018dataset}, which target one type of image).

Not all questions in these datasets are equally useful for training a reasoning model; some are too easy (already trivial for the base model) and some are too hard (unsolvable even with reasoning). Following recent curriculum learning insights~\cite{muennighoff2025s1}, we perform a difficulty-based filtering on both datasets. We prompt three variants of a general multimodal model (Qwen2.5-VL-Instruct with 3B, 7B, 32B parameters) to answer each question 16 times (using nucleus sampling with temperature 1.0). For each question, we record the \emph{pass count}, i.e. the number of trials (out of 16) that produced the correct answer. Figure~\ref{fig:pass-cnt} shows the distribution of pass counts on the text-only m23k and image-based PMC-VQA, for each model size. As model scale increases, more questions achieve high pass counts (e.g. the 32B model answers a larger fraction of questions correctly in a majority of trials). This indicates that the base model’s capability improves with scale, which in turn means that a sufficiently large model can solve many of the questions reliably given enough attempts. For the purposes of training data selection, we focus on medium-difficulty questions that are neither always solved nor hopelessly unsolved. Concretely, we use the results of the 3B model to filter the data: any question with pass count $=0$ (all trials wrong) or $\ge 7$ (correct in at least 7 out of 16 trials) is removed. This retains questions that a smaller model finds neither trivial nor impossible, under the assumption that these medium-difficulty questions will benefit most from reasoning training. After filtering, the text-only dataset is reduced to 16,512 questions and the image-text dataset to 115,456 questions. These filtered datasets are used for all subsequent training of 3B, 7B, and 32B models, ensuring a fair comparison across model scales.

\subsection{Training Strategies}
We train our \ours models on the filtered data under different strategies, as outlined above. We perform SFT and RLVR on the text-only and image-text datasets \emph{separately} to isolate the effect of each data modality. In addition, we experiment with two combined strategies: (a) SFT on text-only data followed by RL on image-text data (denoted SFT\TextCircle + RL\TextImage), and (b) RL on text-only data followed by RL on image-text data (RL\TextCircle + RL\TextImage). Figure~\ref{fig:method}(B) illustrates the training variants. Below, we describe the two core training paradigms in detail:

\paragraph{Supervised Fine-Tuning (SFT).} Supervised fine-tuning forms the foundation of our pipeline. Starting from a general-purpose pretrained multimodal language model (Qwen2.5-VL), we minimize the token-level cross-entropy loss on the curated question-answer pairs (with their reasoning traces). Using teacher-forced learning on the high-quality CoT annotations provides a dense supervision signal, allowing the model to quickly internalize domain-specific medical knowledge, terminology, answer formatting, and the nuanced conventions of clinical explanations. For text-only questions, we use long-form rationales generated by the DeepSeek-R1 model as targets, and for image-based questions, we use GPT-4o-generated rationales. This SFT step teaches the model to emulate the step-by-step reasoning of superior teachers.

\paragraph{Reinforcement Learning with Verifiable Rewards (RLVR).} After SFT, we further refine the model using RL on answer correctness as feedback. We adopt Group Relative Policy Optimization (GRPO), a variant of PPO that operates on a group of sampled outputs. For each question, we sample $N$ reasoning trace rollouts from the model (we use $N=8$ in our experiments). A deterministic verifier then checks each output: if the answer is given in the expected format (e.g., the model produces a chain-of-thought delineated by special tokens and then a final answer choice) and the final answer is correct, a reward $+1$ is assigned; otherwise, a reward $-1$ is assigned. We normalize (whiten) these binary rewards across the group of outputs to obtain advantage estimates. The GRPO algorithm then updates the model policy using a PPO-style clipped objective, where the usual learned value function is replaced by group-based advantage computation. This yields a KL-regularized, contrastive policy update that steadily pushes the model to generate more \emph{verifiably correct reasoning traces} (i.e. reasoning that leads to the correct answer) while constraining it to stay close to the behavior policy (to avoid degeneration). Importantly, RLVR does not require explicit CoT annotations, only a reliable way to verify final answer correctness, making it an appealing method to enhance reasoning using the same data. In our setting, all questions are multiple-choice or otherwise have objectively correct answers, so the reward signal is automatically obtained.

%% file: tabs/comp.tex
\begin{table*}[t]
\centering
\caption{Performance on multimodal medical benchmarks for our baselines. We use greedy decoding to evaluate the ability of the models. \TextCircle means text-only data; \TextImage means image-text data.}
\label{tab:results}
\resizebox{\linewidth}{!}{%
\begin{tabular}{l|ccc|ccc|c} 
\toprule
Model                  & {PMC}              & {MMMU}       & {MedX-M}        & {PathVQA}              & {SLAKE}                & {VQA-Rad}              & {Avg.}                      \\
\midrule
Qwen2.5-VL-3B-Instruct                         & {\cellcolor[rgb]{0.718,0.886,0.804}}44.77 & {\cellcolor[rgb]{0.753,0.902,0.831}}44.12 & {\cellcolor[rgb]{0.718,0.886,0.808}}20.69 & {\cellcolor[rgb]{0.353,0.741,0.549}}61.96 & {\cellcolor[rgb]{0.706,0.882,0.8}}61.30   & {\cellcolor[rgb]{0.616,0.847,0.737}}62.01 & {\cellcolor[rgb]{0.682,0.875,0.784}}49.14  \\ 
\hline
\quad SFT(\TextCircle m23k)                    & 28.53                                     & 32.55                                     & 16.00                                     & 42.74                                     & 43.91                                     & 33.09                                     & 32.80                                      \\ 
\quad SFT(\TextImage PMC)                      & {\cellcolor[rgb]{0.341,0.733,0.541}}54.55 & {\cellcolor[rgb]{0.671,0.867,0.773}}47.84 & {\cellcolor[rgb]{0.671,0.867,0.773}}21.46 & {\cellcolor[rgb]{0.635,0.855,0.749}}52.76 & {\cellcolor[rgb]{0.608,0.843,0.729}}65.79 & {\cellcolor[rgb]{0.69,0.875,0.788}}58.58  & {\cellcolor[rgb]{0.631,0.851,0.745}}50.16  \\ 
\quad SFT(\TextCircle m23k)+RL(\TextImage PMC) & {\cellcolor[rgb]{0.69,0.875,0.788}}46.32  & {\cellcolor[rgb]{0.749,0.898,0.827}}44.31 & {\cellcolor[rgb]{0.729,0.89,0.812}}20.52  & {\cellcolor[rgb]{0.961,0.984,0.973}}43.85 & {\cellcolor[rgb]{0.757,0.902,0.831}}58.49 & {\cellcolor[rgb]{0.784,0.914,0.851}}50.98 & {\cellcolor[rgb]{0.784,0.914,0.851}}44.08  \\ 
\hline
\quad RL(\TextCircle m23k)                     & {\cellcolor[rgb]{0.671,0.867,0.773}}47.32 & {\cellcolor[rgb]{0.341,0.733,0.541}}52.16 & {\cellcolor[rgb]{0.341,0.733,0.541}}22.90 & {\cellcolor[rgb]{0.341,0.733,0.541}}62.28 & {\cellcolor[rgb]{0.671,0.867,0.773}}63.38 & {\cellcolor[rgb]{0.341,0.733,0.541}}71.08 & {\cellcolor[rgb]{0.341,0.733,0.541}}53.19  \\ 
\quad RL(\TextImage PMC)                       & {\cellcolor[rgb]{0.357,0.741,0.553}}54.22 & {\cellcolor[rgb]{0.627,0.851,0.741}}48.43 & {\cellcolor[rgb]{0.663,0.863,0.765}}21.51 & {\cellcolor[rgb]{0.671,0.867,0.773}}51.61 & {\cellcolor[rgb]{0.341,0.733,0.541}}75.56 & {\cellcolor[rgb]{0.608,0.843,0.729}}62.38 & {\cellcolor[rgb]{0.427,0.769,0.604}}52.28  \\ 
\quad RL(\TextCircle m23k)+RL(\TextImage PMC)  & {\cellcolor[rgb]{0.49,0.796,0.647}}51.33  & {\cellcolor[rgb]{0.627,0.851,0.741}}48.43 & {\cellcolor[rgb]{0.412,0.765,0.592}}22.60 & {\cellcolor[rgb]{0.741,0.898,0.824}}49.71 & {\cellcolor[rgb]{0.6,0.839,0.722}}66.11   & {\cellcolor[rgb]{0.671,0.867,0.773}}60.17 & {\cellcolor[rgb]{0.671,0.867,0.773}}49.72  \\ 
\midrule
Qwen2.5-VL-7B-Instruct                         & {\cellcolor[rgb]{0.702,0.878,0.792}}49.30 & {\cellcolor[rgb]{0.561,0.824,0.698}}52.94 & {\cellcolor[rgb]{0.851,0.941,0.898}}18.89 & {\cellcolor[rgb]{0.388,0.753,0.576}}65.39 & {\cellcolor[rgb]{0.671,0.867,0.773}}65.71 & {\cellcolor[rgb]{0.341,0.733,0.541}}68.75 & {\cellcolor[rgb]{0.451,0.78,0.62}}53.50    \\ 
\hline
\quad SFT(\TextCircle m23k)                    & 34.58                                     & 46.86                                     & 16.40                                     & {\cellcolor[rgb]{0.671,0.867,0.773}}56.35 & 54.97                                     & 53.80                                     & 43.83                                      \\ 
\quad SFT(\TextImage PMC)                      & {\cellcolor[rgb]{0.443,0.776,0.612}}54.67 & {\cellcolor[rgb]{0.769,0.906,0.839}}49.80 & {\cellcolor[rgb]{0.698,0.878,0.792}}21.39 & {\cellcolor[rgb]{0.808,0.922,0.867}}53.02 & {\cellcolor[rgb]{0.341,0.733,0.541}}67.71 & {\cellcolor[rgb]{0.702,0.878,0.792}}57.72 & {\cellcolor[rgb]{0.671,0.867,0.773}}50.72  \\ 
\quad SFT(\TextCircle m23k)+RL(\TextImage PMC) & {\cellcolor[rgb]{0.827,0.929,0.878}}43.18 & {\cellcolor[rgb]{0.925,0.969,0.949}}47.84 & {\cellcolor[rgb]{0.671,0.867,0.773}}21.84 & {\cellcolor[rgb]{0.871,0.949,0.914}}51.43 & {\cellcolor[rgb]{0.839,0.937,0.89}}60.34  & {\cellcolor[rgb]{0.898,0.961,0.929}}55.15 & {\cellcolor[rgb]{0.867,0.949,0.91}}46.63   \\ 
\hline
\quad RL(\TextCircle m23k)                     & {\cellcolor[rgb]{0.671,0.867,0.773}}50.67 & {\cellcolor[rgb]{0.341,0.733,0.541}}56.86 & {\cellcolor[rgb]{0.459,0.78,0.624}}24.43  & {\cellcolor[rgb]{0.341,0.733,0.541}}66.83 & {\cellcolor[rgb]{0.659,0.863,0.765}}65.79 & {\cellcolor[rgb]{0.467,0.784,0.631}}64.71 & {\cellcolor[rgb]{0.341,0.733,0.541}}54.88  \\ 
\quad RL(\TextImage PMC)                       & {\cellcolor[rgb]{0.4,0.757,0.584}}55.38   & {\cellcolor[rgb]{0.431,0.773,0.604}}55.29 & {\cellcolor[rgb]{0.482,0.792,0.643}}24.11 & {\cellcolor[rgb]{0.651,0.859,0.757}}57.09 & {\cellcolor[rgb]{0.529,0.812,0.675}}66.59 & {\cellcolor[rgb]{0.506,0.8,0.659}}63.48   & {\cellcolor[rgb]{0.439,0.773,0.612}}53.66  \\ 
\quad RL(\TextCircle m23k)+RL(\TextImage PMC)  & {\cellcolor[rgb]{0.341,0.733,0.541}}56.37 & {\cellcolor[rgb]{0.671,0.867,0.773}}50.98 & {\cellcolor[rgb]{0.341,0.733,0.541}}25.80 & 48.24                                     & {\cellcolor[rgb]{0.875,0.949,0.914}}59.13 & {\cellcolor[rgb]{0.671,0.867,0.773}}58.09 & {\cellcolor[rgb]{0.718,0.886,0.808}}49.77  \\

\bottomrule
\end{tabular}
}
\arrayrulecolor{black}
\end{table*}

%% file: tabs/comp-others.tex
\begin{table*}[t]
\centering
\caption{Performance on multimodal medical benchmarks with other methods. We use greedy decoding to evaluate the ability of the models. \TextCircle means text-only data.
Open Weights (OW): only the model parameters are released;
Open Recipe (OR): data, code, and training details are released, enabling full reproducibility.
}
\label{tab:results-others}
\resizebox{\linewidth}{!}{%
\begin{tabular}{l|c|c|ccc|ccc|c} 
\toprule
Model       & {OW}  &   {OR}         & {PMC}              & {MMMU}       & {MedX-M}        & {PathVQA}              & {SLAKE}                & {VQA-Rad}              & {Avg.}                      \\
\midrule
\multicolumn{10}{c}{General LMM}        \\
\midrule
GPT-4o-mini   & \XSolidBrush& \XSolidBrush        & 51.90                         & 63.53                             & 28.55                              & 63.33                                & 75.24                              & 66.91                                 & 58.24                     \\ 
GPT-4o        & \XSolidBrush& \XSolidBrush            & 58.55                         & 68.82                             & 35.95                              & 72.43                                & 76.44                              & 70.22                                 & 63.74                     \\
\midrule

Gemme 3 4B       & \Checkmark & \XSolidBrush            & 44.42 & 46.67 & 21.89 & 59.24 & 66.59 & 56.86 & 49.28 \\
Gemme 3 27B      & \Checkmark & \XSolidBrush            & 52.05 & 60.78 & 30.80 & 65.70 & 72.60 & 65.20 & 57.86 \\
Qwen2.5-VL-3B-Instruct  & \Checkmark & \XSolidBrush     & 44.77 & 44.12 & 20.69 & 61.96 & 61.30 & 62.01 & 49.14 \\
Qwen2.5-VL-7B-Instruct  & \Checkmark & \XSolidBrush     & 49.30 & 52.94 & 18.89 & 65.39 & 65.71 & 68.75 & 53.50 \\
Qwen2.5-VL-32B-Instruct & \Checkmark & \XSolidBrush     & 53.28 & 63.92 & 27.68 & 67.98 & 73.24 & 75.12 & 60.20 \\
\midrule
\multicolumn{10}{c}{Medical LMM}        \\
\midrule
MedGemma 4B               & \Checkmark & \XSolidBrush & 42.73 & 32.55 & 8.17  & 59.64 & 83.49 & 78.55 & 50.86 \\
MedGemma 27B              & \Checkmark & \XSolidBrush & 36.75 & 35.88 & 12.13 & 62.09 & 77.40 & 72.67 & 49.49 \\
Llava Med v1.5 Mistral 7B & \Checkmark & \Checkmark   & 34.28 & 31.37 & 22.56 & 56.52 & 62.82 & 56.74 & 44.05 \\
HuatuoGPT-Vision-7B       & \Checkmark & \Checkmark   & 53.39 & 50.59 & 22.00 & 63.53 & 75.00 & 63.60 & 54.69 \\
HuatuoGPT-Vision-34B      & \Checkmark & \Checkmark   & 52.54 & 57.06 & 21.80 & 66.72 & 78.85 & 74.26 & 58.54 \\
\midrule
\ours-3B \ \ RL(\TextCircle m23k)    & \Checkmark & \Checkmark          & 47.32 & 52.16 & 22.90 & 62.28 & 63.38 & 71.08 & 53.19  \\ 

\ours-7B \ \ RL(\TextCircle m23k)    & \Checkmark & \Checkmark          & 50.67  & 56.86 & 24.43 & 66.83 & 65.79 & 64.71 & 54.88  \\ 

\ours-32B RL(\TextCircle m23k)   & \Checkmark & \Checkmark     &54.37 & 70.00 & 34.60 & 68.82 & 73.96 & 76.96 & 63.12   \\ 

\bottomrule
\end{tabular}
}
\arrayrulecolor{black}
\end{table*}


%% file: sections/4_experiments.tex
\section{Experiments}
\subsection{Implementation Details}
We initialize our models from the Qwen2.5-VL checkpoint. 
For SFT, we fine-tune the model for 3 epochs with a batch size of 32 and learning rate $1\times10^{-4}$. 
For RLVR, we train using GRPO for 5 epochs on the text-only data and 1 epoch on the image-text data, with a learning rate of $1\times10^{-6}$. We set the total batch size to 128 for text-only RL (sufficient to sample 8 rollouts per question) and 256 for image-text RL (since each sample includes image features). For experiments where RL is continued on a second dataset (e.g., applying RL on PMC-VQA after an SFT on m23k), we reduce the batch size (to 64) during the second stage to accommodate the longer sequence lengths (the combined image+CoT+answer sequence can reach $\sim$2048 tokens). All models are trained on 8$\times$H100 GPUs using mixed precision, except the 32B model, which is trained on 32 GPUs.

\subsection{Evaluation}
We evaluate our models on a suite of \textbf{six} multimodal medical QA benchmarks, which can be divided into two categories: (1) \textit{general-domain medical QA} and (2) \textit{modality-specific QA}. The general-domain evaluations include the test set of PMC-VQA~\cite{zhang2023pmc} (for direct comparison, since our models train on a filtered subset of its training data), the validation set of MMMU-Health~\cite{yue2024mmmu} (the health and medicine portion of the MMMU benchmark), and MedXpert-MM~\cite{zuo2025medxpertqa}, a challenging benchmark requiring complex reasoning over multimodal inputs. The modality-specific evaluations include PathVQA~\cite{he2020pathvqa} (pathology images), SLAKE~\cite{liu2021slake} (slit-lamp ophthalmology images) and VQA-Rad~\cite{lau2018dataset} (radiology X-rays). Together, these six datasets cover a broad range of medical visual question answering scenarios, from generic biomedical knowledge to highly specialized imaging tasks.

For each benchmark, we report the accuracy (\% of questions answered correctly). Model responses are generated using greedy decoding (temperature $0$) to evaluate base capability without sampling variance. We note that even with deterministic decoding, slight nondeterminism in the inference engine (due to floating-point precision) can cause minimal variability; thus, we run each evaluation 3 times and report the average accuracy (the standard deviation was below 0.1 and is provided in the appendix for completeness). In the result tables, we use the notation \TextCircle to indicate models trained on the text-only (m23k) data and \TextImage for models trained on the image-text (PMC-VQA) data. For example, “SFT\TextCircle” denotes a model fine-tuned on text-only CoT data, and “RL\TextCircle+RL\TextImage” denotes a model first trained with RL on text-only data then further with RL on image-text data.

\subsection{Results}
\paragraph{Impact of Training Paradigm (SFT vs. RLVR).} Table~\ref{tab:results} summarizes the performance of the Qwen2.5-VL 3B and 7B models under various training recipes. We observe that RLVR-trained models consistently outperform SFT-trained models of the same size across all benchmarks. For the 3B base, RLVR on text-only data (RL\TextCircle) achieves 53.19\% average accuracy, versus 32.80\% for SFT on text-only (SFT\TextCircle) (a dramatic drop below the 49.14\% base performance). Similarly, the 7B RL\TextCircle model reaches 54.88\% average, compared to 43.83\% for SFT\TextCircle (again, SFT underperforms even the 53.50\% base model). These results confirm that simply fine-tuning on distilled CoT data does \emph{not} guarantee better performance – in fact, it may overload the model with long, possibly mismatched rationales that hurt its effectiveness on multimodal QA. In contrast, RLVR directly optimizes the model’s own reasoning policy and proves markedly more effective at improving accuracy.

\paragraph{Impact of Training Data (Text-only vs. Image-text).} The choice of training data modality also has a significant effect. From Table~\ref{tab:results}, training on the text-only data tends to yield better results than training on the image-text data. For instance, the 7B RL\TextCircle (54.88\% avg) outperforms RL\TextImage (53.66\% avg). 
However, SFT on the text-only CoT data consistently \textit{harms} performs relative to the base model (43.83\% for 7B SFT\TextCircle), whereas SFT on the multimodal data yields a slight improvement over base on some benchmarks (e.g., +1-2\% on PathVQA, SLAKE) but overall comparable average (50.72\% SFT\TextImage vs 53.50\% base). 
We hypothesize that the long, text-only rationales distilled from a text-based LRM (DeepSeek) may not align well with the needs of a multimodal model that also has to interpret images. 
The image-based data, while noisy, at least engages the model’s visual processing during training, which might explain why SFT\TextImage does not drastically degrade performance. 
Nonetheless, the strongest gains come from RLVR on text-only data, which boosts performance substantially (e.g., +4.05\% for 3B, +1.38\% for 7B, compared to base). 
RLVR on the multimodal data also improves over base, but to a lesser degree.  These results highlight that in RLVR,  high-quality \emph{textual} QA data (with verifiable answers) can be more valuable than larger but noisier image-based data for training the reasoning capability of multimodal models. Improving the quality of multimodal training data remains an important challenge (see Discussion).

\paragraph{Combined Training Strategies.} We also evaluated whether combining text-only and image-text training yields further benefits. Two combinations were tried: SFT on text then RL on images (SFT\TextCircle+RL\TextImage), and RL on text then RL on images (RL\TextCircle+RL\TextImage). As Table~\ref{tab:results} shows, neither strategy provided gains over the single-modality RL training. In fact, for the 7B model, SFT\TextCircle+RL\TextImage (53.07\% avg) was worse than RL\TextImage alone (53.66\%), and RL\TextCircle+RL\TextImage (49.77\%) fell behind RL\TextCircle (54.88\%). For the 3B model, similar results are observed. 
It appears that after a model has been optimized on the text-only data, adding the image-text data (even via RL) can hinder the reasoning capability, resulting in a net drop in performance.
We conclude that the best recipe in our study is to apply RLVR directly on a high-quality text-only reasoning dataset. This produces the top results for both 3B and 7B. In most cases, adding an SFT stage or an extra RL stage on image data does not help, and in the worst case, it reduces accuracy.

\paragraph{Effect of Model Scale.} Increasing the model size clearly improves performance across the board. The 7B models outperform the 3B models in every corresponding setting (comparing rows in Table~\ref{tab:results}). For example, the base 7B is 4.36\% higher on average than base 3B; the RL\TextCircle 7B is +1.69\% higher than its counterpart 3B; and SFT\TextImage 7B is +0.56\% higher than SFT\TextImage 3B. On certain benchmarks like MedXpert-MM (which is especially challenging and requires complex reasoning), the gap is more pronounced: the best 7B (RL\TextCircle) attains 24.43\% versus 22.90\% for the best 3B, and 7B SFT\TextCircle achieves 16.40\% vs 16.00\% for 3B (both quite low). This trend suggests that larger models have more capacity to learn medical knowledge and to benefit from the reasoning training. Pushing to even larger scales may continue to yield gains (we test a 32B model below).

\paragraph{Comparison to Previous Models.} In Table~\ref{tab:results-others}, we compare our \ours models against prior open-source medical VLMs and against GPT-4-based models. Our 7B RLVR-trained model achieves an average score of 54.88\%, which is 3–4\% higher than the reported performance of HuatuoGPT-Vision-7B-Qwen2.5 (48.60\% avg) and also above LLaVA-Med v1.5 (Mistral-7B).
On general-domain benchmarks like MedXpert-MM, our advantage is even larger: \ours-7B scores 24.43\% vs HuatuoGPT-Vision’s 22.00\%. This demonstrates the benefit of our focused reasoning training. HuatuoGPT-Vision was primarily trained with instruction tuning on multimodal data (and a bit of RLHF), and it underperforms on challenging reasoning questions. We also note that HuatuoGPT-Vision reportedly suffered a large performance drop on generic medical QA after its multimodal fine-tuning (similar to our observation that SFT on image data can hurt general QA). 
In contrast, our RLVR approach improved performance without such trade-offs. Finally, our \ours-32B (RL on text-only) reaches 63.12\% average accuracy, surpassing the GPT-4o-mini model (58.24\%) and essentially matching the full GPT-4o (63.74\%) on these benchmarks. This is a notable result: it suggests that with sufficient model size and proper training, open models can approach the performance of proprietary models like GPT-4 on specialized tasks. We emphasize that our entire training pipeline, data, and models are open-source, providing a foundation for the community to build upon.

\paragraph{Qualitative Results.} We provide a few anecdotal examples of our model’s outputs in Figure~\ref{fig:sample} to illustrate the reasoning quality of text-only RLVR training. More qualitative results of 3B, 7B, and 32B models can be found in the supplemental materials.

%% file: sections/5_discussions.tex
\section{Conclusion}
In this work, we presented \ours, a set of baseline multimodal medical reasoning models built by combining large vision-language models with advanced reasoning training paradigms. We carried out a systematic study of supervised CoT fine-tuning versus reinforcement learning (GRPO-based RLVR) for teaching a multimodal model to reason about medical questions. Our experiments show that RLVR is markedly more effective than CoT fine-tuning in improving model performance, especially when using high-quality text-only medical QA data. We also found that models trained on text-only data generalize better than those trained on image-text data, highlighting a data quality issue in current multimodal corpora. By training models at multiple scales, we demonstrated a clear benefit to larger model size: our 7B \ours achieves state-of-the-art results among open models on six benchmarks, and a 32B variant reaches parity with a GPT-4-based competitor. Our work provides not only strong baseline models for the community but also insights into training strategies for multimodal reasoning. In future work, we plan to address the limitations identified (data quality, curriculum, broader tasks) and hope that \ours will inspire further research in reliable and transparent medical AI.

%% file: sections/6_supp.tex
\label{sec:appendix}

\begin{figure*}[t]
  \centering

  \subfigure[\small{Qwen2.5-VL-In-3B m23k}]{
    \includegraphics[width=0.3\textwidth]{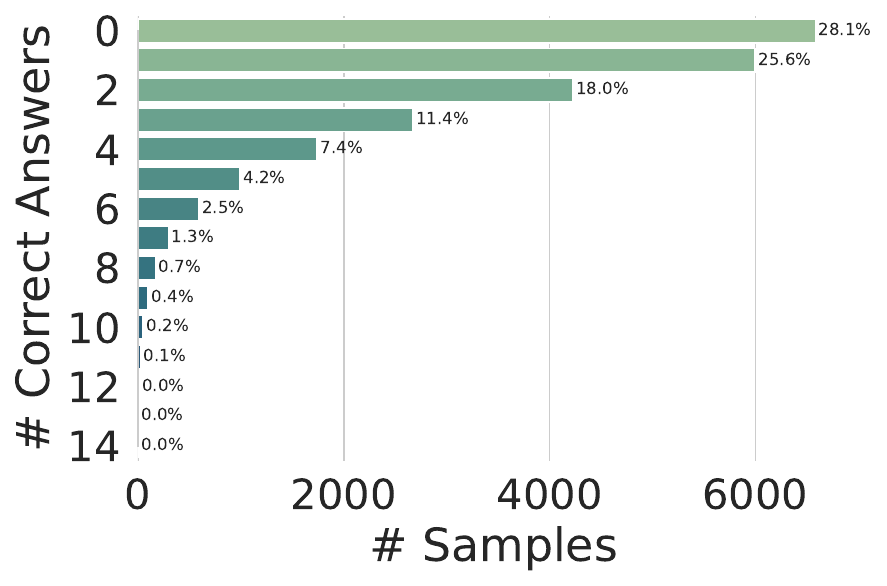}
  }\hfill
  \subfigure[\small{Qwen2.5-VL-In-7B m23k}]{
    \includegraphics[width=0.3\textwidth]{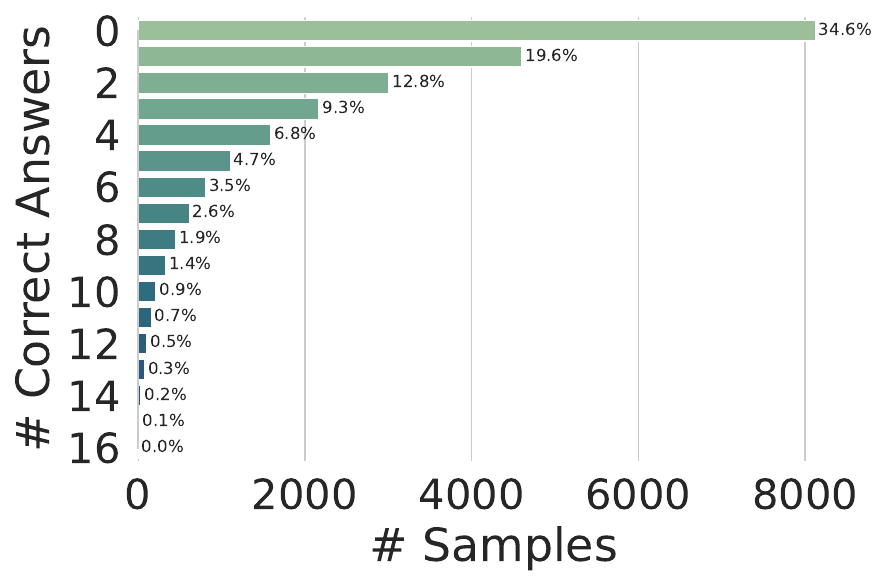}
  }\hfill
  \subfigure[\small{Qwen2.5-VL-In-32B m23k}]{
    \includegraphics[width=0.3\textwidth]{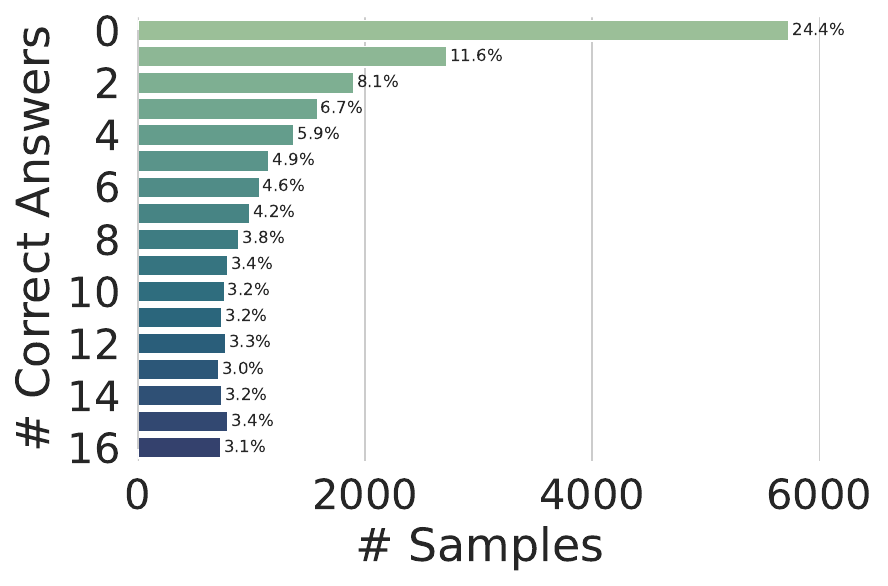}
  }

  \vspace{0.6em}

  \subfigure[\small{Qwen2.5-VL-In-3B PMC-VQA}]{
    \includegraphics[width=0.3\textwidth]{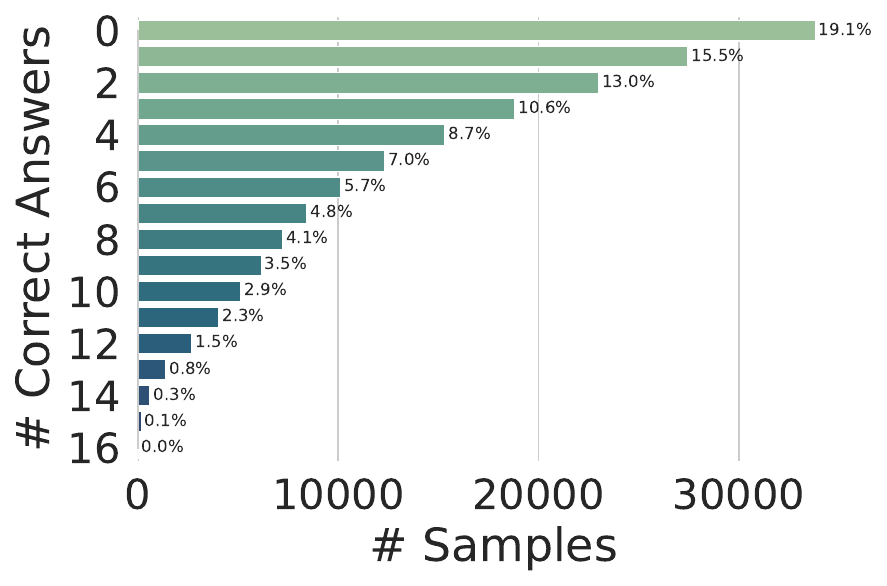}
  }\hfill
  \subfigure[\small{Qwen2.5-VL-In-7B PMC-VQA}]{
    \includegraphics[width=0.3\textwidth]{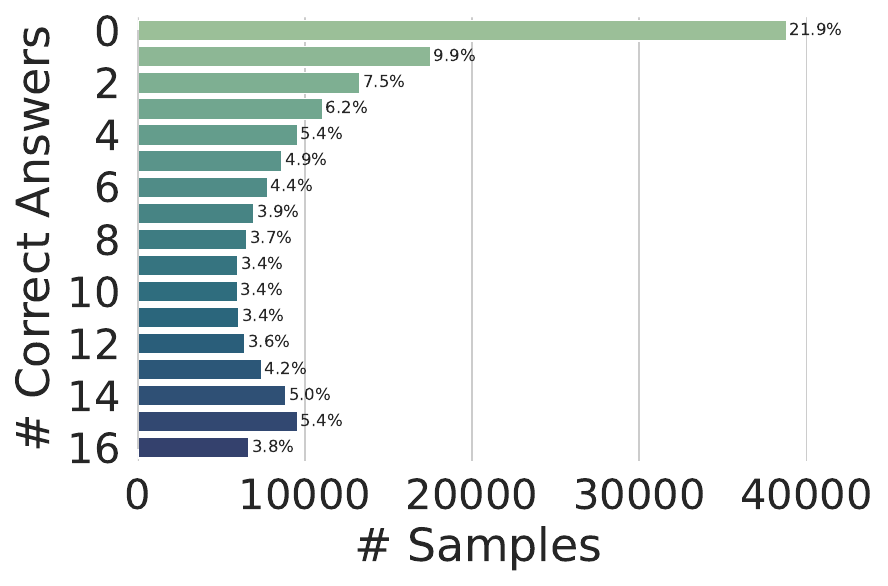}
  }\hfill
  \subfigure[\small{Qwen2.5-VL-In-32B PMC-VQA}]{
    \includegraphics[width=0.3\textwidth]{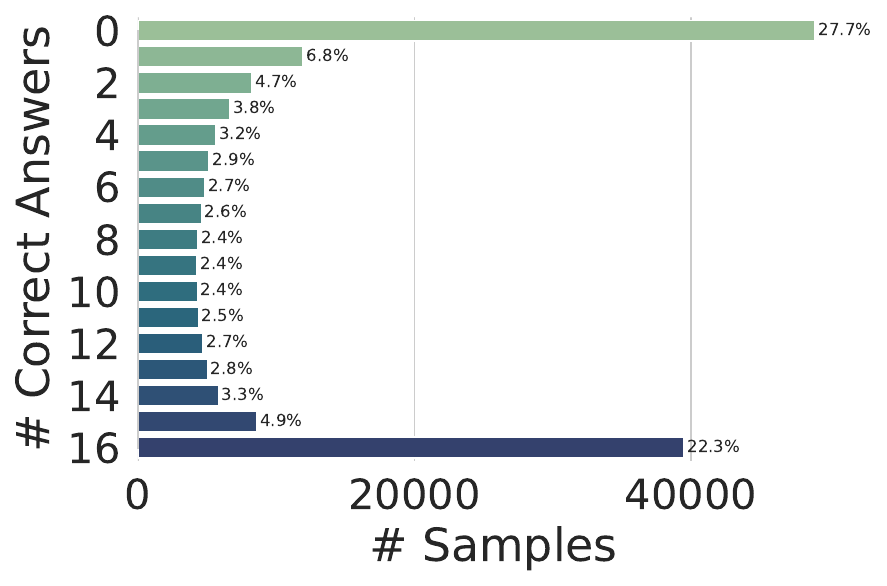}
  }

    \caption{
        Probing the questions difficulty with Qwen2.5-VL-Instruct. For each question, we generate 16 answers. 
        Then we draw the pie plots for the pass count. When the scale of the multimodal LLM increased, the number of high pass count questions increased. 
        This indicate the potential of the models, especially for latter RLVR training, which encourage the models improve this possibility to answer questions correctly.
        The pass count are used for latter data filtering.
    }
    \label{fig:pass-cnt}
\end{figure*}

\section{Implementation Details}

We implement our training pipeline using two stages: supervised fine-tuning (SFT) followed by reinforcement learning with verifiable rewards (RLVR). For SFT, we employ distributed training using PyTorch's torchrun with FSDP (Fully Sharded Data Parallel) configuration across 8 GPUs per node. The SFT stage uses a learning rate of 1e-5 with cosine scheduler, warmup ratio of 0.05, weight decay of 1e-4, and trains for 5 epochs with a global batch size of 16. We utilize gradient checkpointing and bf16 precision to optimize memory usage. For the RLVR stage, we use the VERL framework with GRPO (Group Relative Policy Optimization) as the advantage estimator. The RL training employs a smaller learning rate of 1e-6, KL divergence regularization with coefficient 0.01, and generates 8 samples per prompt during rollout. We implement a custom reward function that combines format adherence (ensuring responses follow the \verb|<think>...</think> <answer>...</answer>| structure) with accuracy rewards based on exact answer matching. The training uses VLLM for efficient inference during rollout generation with tensor model parallelism across 2 GPUs and 60\% GPU memory utilization. Throughout both stages, we use the instruction prompt template below.

\begin{AIbox}{Prompt Template}
\begin{verbatim}
You will solve a problem/request. 
You should provide your thoughts 
within <think> </think> tags before 
providing the answer.\nWrite your 
final answer within <answer> </answer> 
tags.\n{{ Question }}\n{{ Options }}
\end{verbatim}
\end{AIbox}

\section{Qualitative Results}

\begin{figure*}[t]
    \centering
    \includegraphics[width=0.93\linewidth]{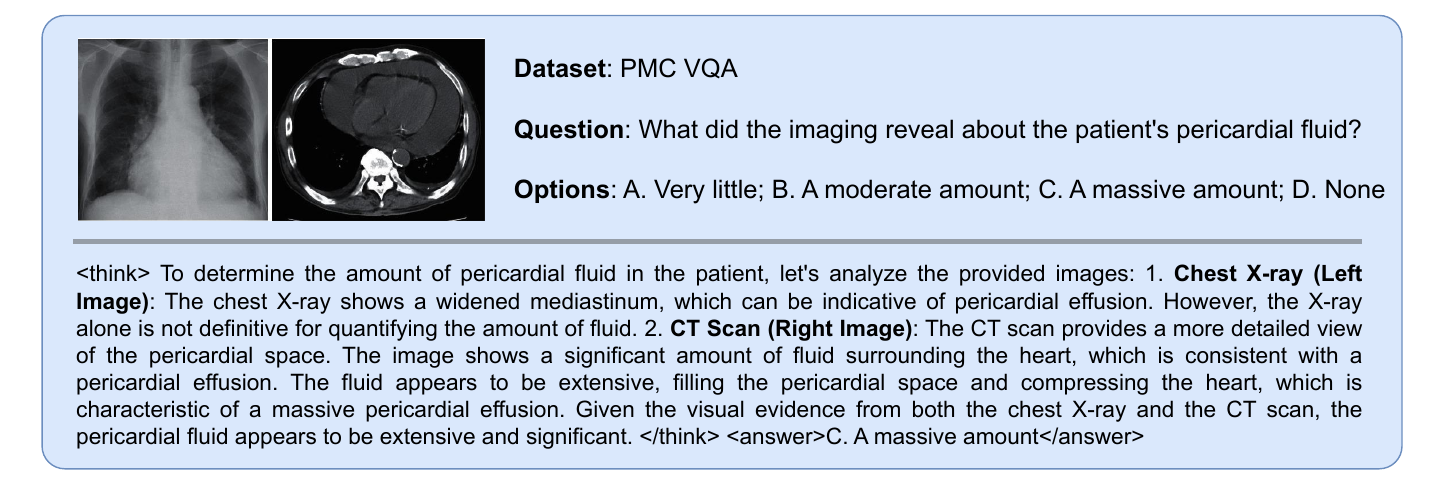}
    \includegraphics[width=0.93\linewidth]{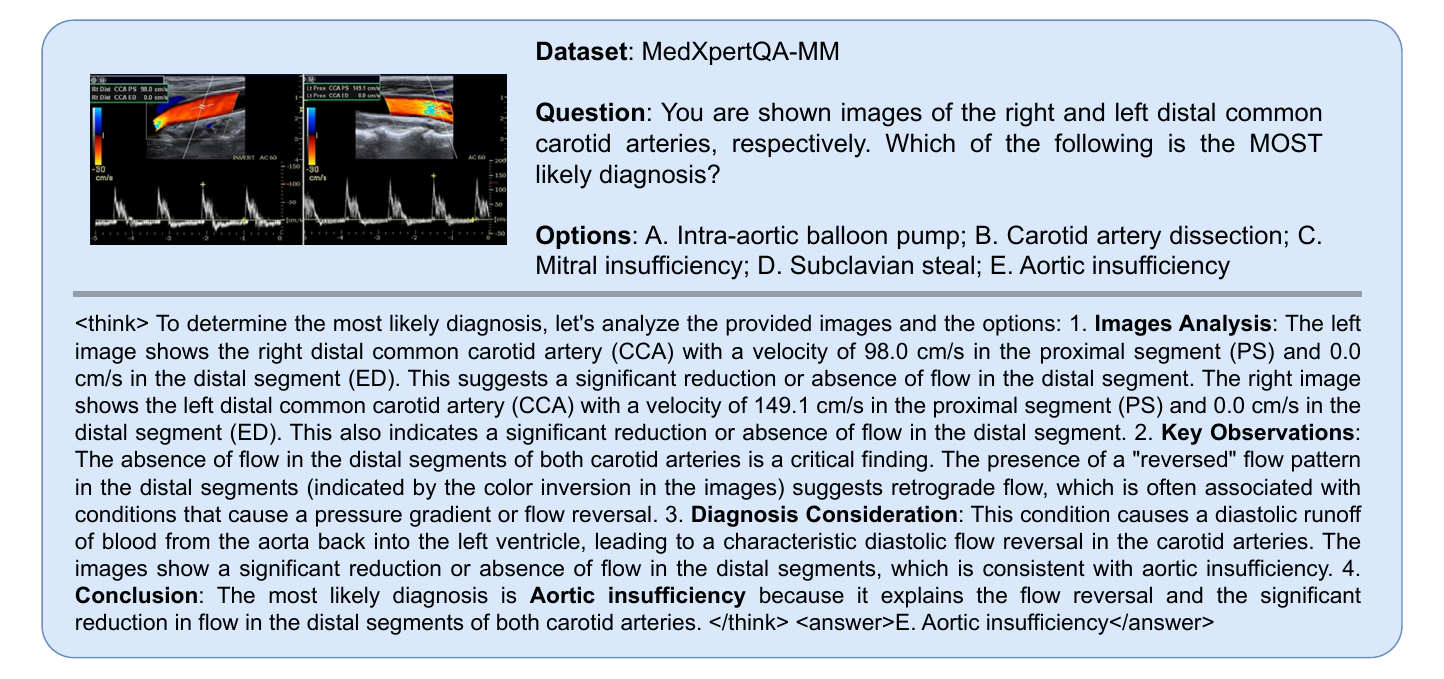}
    \includegraphics[width=0.93\linewidth]{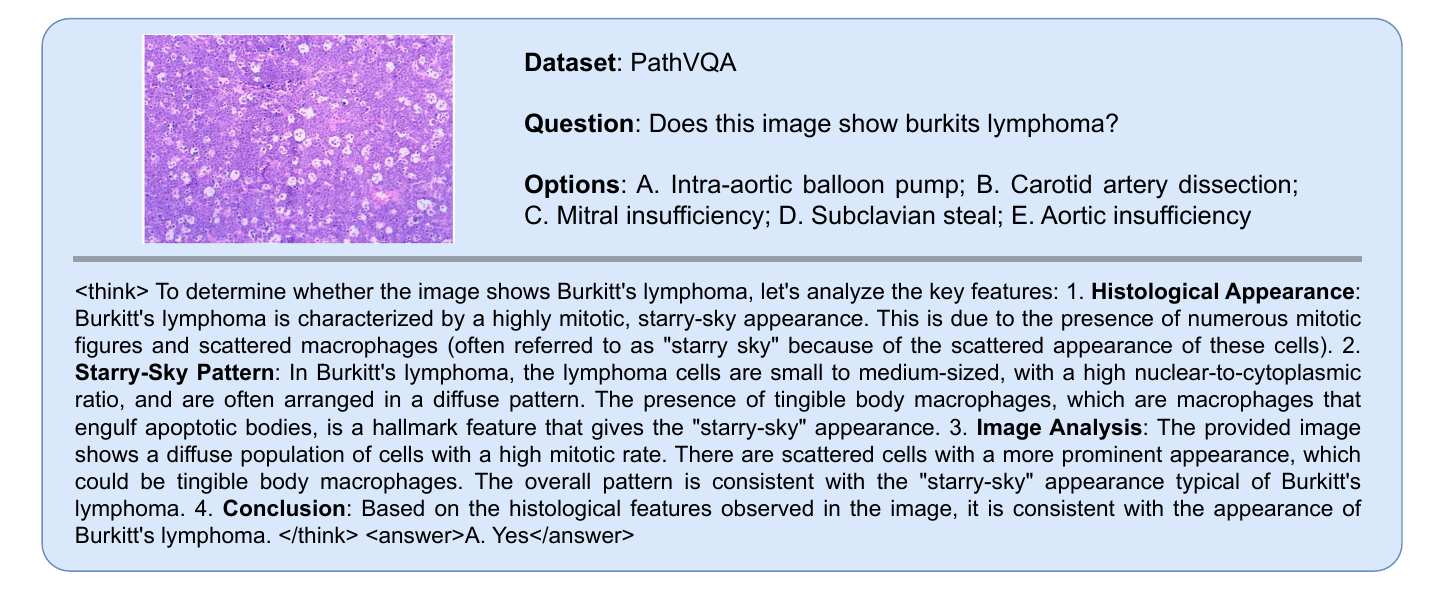}
    \caption{Case study on multiple medical VQA benchmarks with our 32B text-only RLVR model. Out \ours demonstrates robust reasoning capability across various imaging modalities.}
    \label{fig:sample}
\end{figure*}


More qualitative results (text-only RLVR 3B, 7B, and 32B) can be found in the \textbf{supplemental material}.


\section{Discussions}
\paragraph{Quality of Training Data.} One striking observation is the performance gap between models trained on text-only data versus image-text data. In our experiments, models trained purely on the PMC-VQA image-text corpus often \emph{lost} exhibit capability relative to their starting point (especially for general QA tasks), whereas models trained on the m23k text-only corpus made clear gains. We suspect the primary cause is the quality of the training data. The PMC-VQA dataset was generated automatically by GPT-3.5 from journal figures and captions. Many of the questions may be simplistic or flawed, and the answers might not always require deep reasoning (or could even be incorrect in some cases). In contrast, the text-only m23k dataset is derived from human-authored exam questions and has higher factual and linguistic quality, supplemented by expert-generated reasoning chains. Unfortunately, besides PMC-VQA, other “general” multimodal medical QA datasets compiled from various sources (e.g., OmniMedQA~\cite{hu2024omnimedvqa}, GMAI-Bench~\cite{ye2024gmai}) currently do not contain training splits and thus are not directly usable for model training. This highlights the need for better multimodal medical QA data. An encouraging direction is the emergence of high-quality, human-curated medical image report datasets (radiology reports, pathology reports, etc.), which could be leveraged to generate more realistic multimodal QA pairs. In future work, we plan to incorporate such sources to improve the image-text training signal.

\paragraph{Difficulty-Based Filtering.} In applying RLVR, it is important to present the model with training examples of appropriate difficulty. We performed a simple filtering by removing questions that a smaller model got either 0/16 or $>$7/16 correct. This “one-size-fits-all” filter was then applied uniformly for training all model sizes. In reality, different model scales have different ability levels, and an optimal curriculum might adjust the filtering threshold per model (a form of capacity-aware data selection). More adaptive curriculum learning strategies could further improve RLVR training by continuously calibrating question difficulty to the model’s growing competence. We leave a full exploration of curriculum learning for medical reasoning to future work.

\paragraph{Effectiveness of RLVR in the Medical Domain.} Our results validate that RLVR is a powerful approach for improving reasoning in medical QA, consistent with findings in other domains. With only a few epochs of RL (a relatively small compute budget compared to pretraining), we observed significant gains in the model’s ability to arrive at correct answers. This improvement can be interpreted as an increase in \textit{sampling efficiency}: after RLVR, the model is far more likely to produce a correct answer in a single try, whereas the base model might need multiple attempts (as illustrated by Figure~\ref{fig:pass-cnt}). Of course, RLVR is not a magic bullet; its success still depends on the diversity and difficulty of the training questions and the reliability of the reward signal. In our case, we used exact-match answer checking, which is straightforward for multiple-choice questions. Extending RLVR to open-ended generation or multi-step clinical reasoning (where reward shaping is harder) is an interesting challenge. We believe scaling up the RLVR approach, with larger verification datasets and more compute, could yield even stronger medical reasoners. This work provides a first step in that direction using openly available tools.

\paragraph{Task Scope and Future Work.} So far, we have limited training and evaluation to single-turn question-answering tasks. However, real-world medical scenarios often involve more interactive and diverse tasks: multi-turn dialogues with patients, grounding textual descriptions in images (e.g., locating findings in an image), and reasoning about temporal sequences of images or data. Our current \ours could potentially be adapted to some of these tasks, but we have not specifically trained or tested it on them. In the future, we aim to extend our framework beyond QA format, incorporating vision-language grounding tasks, dialogue agents for patient interaction, and other medically relevant capabilities. We are hopeful that the combination of explicit reasoning and multimodal understanding, as demonstrated in \ours, can serve as a foundation for more advanced clinical AI systems.

\section{Limitations}

While \ours aims to advance the state of multimodal medical QA, there are several limitations to acknowledge. First, the quality of the image-text training data (PMC-VQA) is a concern. Because this dataset was synthetically generated by MLLMs, it likely contains noise and biases, which in turn limited the gains from training on it. Future improvements will require either cleaning this data or obtaining higher-quality human-curated multimodal data. Second, our difficulty-based filtering strategy was static and based on the 3B model’s performance. This may not have been optimal for the larger models; a more dynamic or model-specific curriculum could yield better results. Third, our RLVR training was relatively short and task-specific, leveraging only QA-format rewards. As a result, the models are specialized to the QA task format and may not generalize to other interactions (e.g., multi-turn conversations or explanatory responses) without additional fine-tuning. Finally, although we tested a 32B model, even larger models or more diverse pretraining might be needed to capture the full complexity of real-world medical reasoning. We release our models and code in hopes that the community can build on them to address these limitations.